\documentclass{article}

\usepackage{arxiv}

\usepackage[utf8]{inputenc} 
\usepackage[T1]{fontenc}    
\usepackage{hyperref}       
\usepackage{url}            
\usepackage{booktabs}       
\usepackage{amsfonts}       
\usepackage{amsmath}
\usepackage{nicefrac}       
\usepackage{microtype}      
\usepackage{lipsum}
\usepackage{enumitem}
\usepackage{graphicx}
\usepackage{tikz}
\usetikzlibrary{arrows.meta,positioning,calc,shapes.geometric}
\usepackage[utf8]{inputenc}
\usepackage{abstract}

\usepackage{amsmath,amssymb}
\usepackage{booktabs}
\usepackage{tabularx}
\usepackage{graphicx}
\usepackage{subcaption}
\usepackage{listings}
\usepackage{xcolor}
\usepackage{hyperref}
\usepackage{enumitem}
\usepackage{caption}
\usepackage{multirow}
\usepackage{tikz}
\usetikzlibrary{shapes.geometric,arrows.meta,positioning,fit,backgrounds,calc,matrix}

\hypersetup{
	colorlinks=true,
	linkcolor=blue!50!black,
	urlcolor=blue!60!black,
	citecolor=blue!50!black
}

\definecolor{ffcasebg}{RGB}{248,248,250}
\definecolor{ffheader}{RGB}{30,80,140}
\definecolor{ffanchor}{RGB}{40,140,80}
\definecolor{ffactive}{RGB}{200,90,30}

\lstdefinestyle{shell}{
	basicstyle=\footnotesize\ttfamily,
	backgroundcolor=\color{gray!8},
	frame=single,framerule=0pt,
	breaklines=true,columns=fullflexible,
	aboveskip=6pt,belowskip=6pt
}
\graphicspath{ {./images/} }

\title{AutoFOAM: The Self-Refining Autonomous OpenFOAM Agent}

\author{
  Arun Govind Neelan$^{1}$ \\
  $^{1}$SimuNetics, Kanyakumari, Tamil Nadu -- 629173 \\
  \texttt{arungovindneelan@gmail.com}
  \And
  A Seshaditya$^{2,3}$ \\
  $^{2}$Onnes Cryogenics, Hyderabad, India \\
  $^{3}$Quasi AI, Berlin, Germany \\
  \texttt{adi@onnes.in} \\
  \texttt{aditya@quasi.digital}
}

\begin{document}
\maketitle
\begin{abstract}
Computational Fluid Dynamics (CFD) is an indispensable tool in modern engineering, yet harnessing robust open-source solvers like OpenFOAM requires extensive domain expertise, steep learning curves, and meticulous manual configuration. To eliminate this technical barrier, we present \textbf{AutoFOAM}, a self-evolving large-language-model (LLM) agent that autonomously authors, executes, scores, and iteratively improves OpenFOAM cases directly from natural-language prompts. Powered by a 14-billion-parameter Qwen-chat backbone, the model is initially fine-tuned on 252 prompts covering seven deterministically selected OpenFOAM solvers, thirteen parametric mesh templates, and a structured $y^+$-aware numerical policy. The system's core innovation is a rigorous seven-layer evolution loop that automates the transition from text to syntactically correct C++ dictionaries. 
To prevent model degeneration under repeated self-training, the agent employs three complementary anti-collapse streams: RAG-augmented retry context, surgical dictionary-level patching, and prompt-diversity paraphrasing. By bridging generative artificial intelligence with rigorous fluid simulations, \textbf{AutoFOAM} accelerates rapid prototyping and democratizes advanced, self-correcting CFD workflows.
\end{abstract}


\section{Introduction}
\label{sec:introduction}

Computational Fluid Dynamics (CFD) is an essential pillar of modern engineering, driving design and analysis across aerospace, automotive, environmental, and biomedical industries \cite{weller1998tensor, anderson2020computational}. Among the available tools, OpenFOAM stands out as the premier open-source simulation framework, offering unparalleled flexibility and access to the underlying physics \cite{openfoam_foundation}. However, this flexibility comes at a significant cost: OpenFOAM requires users to navigate a steep learning curve, manually configuring dozens of interdependent, syntactically rigid C++ dictionaries to define meshes, boundary conditions, physical properties, and numerical schemes. For non-specialists, the transition from a high-level physical concept to an executable simulation is often a brittle, error-prone process.

Recently, Large Language Models (LLMs) have demonstrated remarkable proficiency in code generation and logical reasoning \cite{brown2020language, touvron2023llama}. Yet, applying general-purpose LLMs off-the-shelf to CFD workflows typically yields poor results. Standard models frequently hallucinate unsupported OpenFOAM keywords, fail to maintain consistency across boundary condition files (e.g., mismatching \texttt{U} and \texttt{p} boundary types), and lack the domain physics intuition required for stable numerical setups, such as enforcing correct $y^+$ constraints for turbulence modeling. 
To overcome these fundamental barriers, we present \textbf{AutoFOAM}, a fully autonomous, self-evolving LLM agent capable of authoring, executing, scoring, and systematically improving OpenFOAM cases strictly from natural-language prompts. Rather than relying on a static set of rules or zero-shot prompting of closed-source models, AutoFOAM introduces a robust self-improvement architecture built upon a 14-billion-parameter Qwen-2.5 Model \cite{hui2024qwen2, bai2023qwen}. The model is initially fine-tuned on a deterministically curated dataset of 252 prompts, which systematically cover seven OpenFOAM solver families, thirteen parametric mesh templates, and a strictly enforced $y^+$-aware numerical policy.

The core innovation of this work is the development of a \emph{seven-layer evolution loop} designed to safely scale the agent's capabilities without human intervention. This loop tightly integrates real-time log parsing and solver self-correction with advanced reinforcement learning techniques, specifically Direct Preference Optimization (DPO) applied to trajectory retry pairs \cite{rafailov2023direct}. To ensure long-term learning stability, we introduce three complementary anti-collapse streams: Retrieval-Augmented Generation (RAG), retry context, surgical dictionary-level patching, and prompt-diversity paraphrasing, which collectively prevent the agent from degenerating during continuous self-training.
 We observe marked improvements across solver-family match rates, mean reward scores, and first-pass simulation success.
 
 Recent initiatives such as the development of FoamGPT \cite{yue2025foamgpt}, the retrieval-augmented architecture of OpenFOAMGPT \cite{pandey2025openfoamgpt}, and the multi-agent frameworks of Foam-Agent 2.0 \cite{yue2025foamagent2}—highlight a rapidly growing momentum toward integrating Large Language Models (LLMs) into the physical sciences to automate computational workflows. Comprehensive benchmarking suites like CFDLLMBench \cite{somasekharan2025cfdllmbench} further validate the necessity of these autonomous systems.

In summary, the main contributions of this paper are:
\begin{itemize}
    \item \textbf{An Autonomous CFD Agent Workflow:} We detail the architecture of AutoFOAM, capable of translating natural language into syntactically correct, physics-aware OpenFOAM dictionaries.
    \item \textbf{The Seven-Layer Evolution Loop:} We introduce a novel training pipeline combining in-run self-correction, automatic dataset curation, Supervised Fine-Tuning (SFT), DPO on retry pairs, anchor mixing, active learning, and a strict regression gate to foster continuous, monotonic model improvement.
    \item \textbf{Anti-Collapse Mechanisms for Physical Simulators:} We propose and validate three distinct streams (RAG-augmented context, dictionary patching, and paraphrasing) that successfully mitigate model degeneration during self-play on rigid simulation tasks.
\end{itemize}

The work is organized to reflect the progression from foundational architecture to autonomous deployment. Section \ref{sec:arch} details the AutoFOAM framework, delineating the deterministic execution boundaries and the multi-objective reward formulation. Section \ref{sec:dataset} outlines our foundational data curation and dynamic trajectory mining strategies. Section \ref{sec:evolve} formally defines the seven-layer self-evolution pipeline designed to prevent model collapse. In Section \ref{sec:results}, we validate the system against a rigorous out-of-distribution benchmark, quantifying both zero-shot physical reasoning and pipeline stability. Finally, Section \ref{sec:limitations} discusses the current operational envelope, and Section \ref{sec:conclusion} synthesizes our findings.

\section{System Architecture and Methodology}
\label{sec:arch}

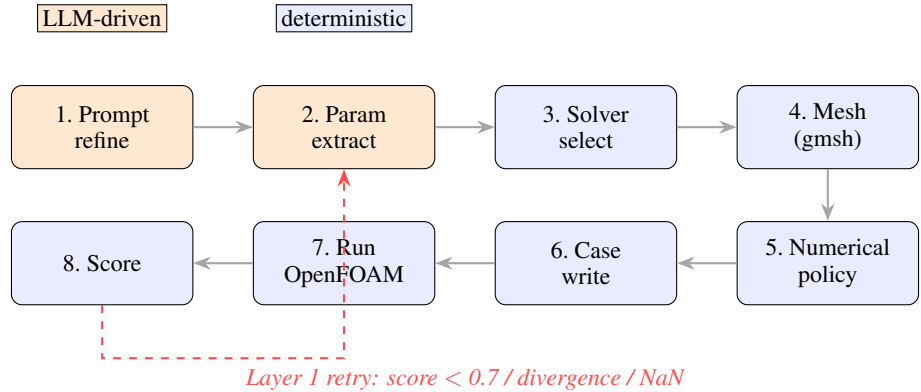
\begin{figure}[h]
    \centering
    \begin{tikzpicture}[
        every node/.style={font=\small},
        stage/.style={rectangle, rounded corners, draw, minimum width=24mm,
            minimum height=11mm, align=center, fill=blue!8,
            inner sep=2pt},
        llmstage/.style={rectangle, rounded corners, draw, minimum width=24mm,
            minimum height=11mm, align=center, fill=orange!18,
            inner sep=2pt},
        arrow/.style={-{Stealth[scale=1.0]}, thick, gray!70},
        x=32mm, y=18mm,
        ]
        
        \node[llmstage] (n1) at (0,1) {1.~Prompt\\refine};
        \node[llmstage] (n2) at (1,1) {2.~Param\\extract};
        \node[stage]    (n3) at (2,1) {3.~Solver\\select};
        \node[stage]    (n4) at (3,1) {4.~Mesh\\(gmsh)};
        
        \node[stage]    (n5) at (3,0) {5.~Numerical\\policy};
        \node[stage]    (n6) at (2,0) {6.~Case\\write};
        \node[stage]    (n7) at (1,0) {7.~Run\\OpenFOAM};
        \node[stage]    (n8) at (0,0) {8.~Score};
        
        \draw[arrow] (n1) -- (n2);
        \draw[arrow] (n2) -- (n3);
        \draw[arrow] (n3) -- (n4);
        \draw[arrow] (n4) -- (n5);
        \draw[arrow] (n5) -- (n6);
        \draw[arrow] (n6) -- (n7);
        \draw[arrow] (n7) -- (n8);
        
        \draw[arrow, dashed, red!70]
        (n8.south) -- ++(0,-7mm)
        -| ($(n2.south)+(0,-7mm)$) -- (n2.south);
        
        \node[font=\footnotesize\itshape, red!70] at (1.5,-0.85)
        {Layer 1 retry: score $<$ 0.7 / divergence / NaN};
        
        \node[draw, fill=orange!18, inner sep=2pt, font=\footnotesize] at (0,1.8) (lega) {LLM-driven};
        \node[draw, fill=blue!8,    inner sep=2pt, font=\footnotesize] at (1,1.8) {deterministic};
        
    \end{tikzpicture}
    \caption{Eight-stage agent pipeline. The LLM-driven stages (orange) manage semantic interpretation and context parsing, while the deterministic stages (blue) handle execution and physical validation. }
    \label{fig:pipeline}
\end{figure}

To effectively control the search space during generation, the proposed agent architecture is explicitly modeled as an eight-step composite structure (Figure~\ref{fig:pipeline}).  The Large Language Model (LLM) is called only at three key decision points: prompt refinement, parameter extraction, and retry generation. The remaining deterministic steps – routing, meshing, dictionary generation, solving, and scoring – are dictated by strict logic.

\paragraph{Constrained Parameter Extraction.}
Semantic reasoning is powered by vLLM \cite{vllm} to ensure high-throughput inference. To strictly enforce valid output topologies and prevent structural hallucinations, the parameter extraction stage wraps generation within \texttt{xgrammar} \cite{xgrammar} JSON-schema constraints. This forces the model to emit a standardized \texttt{CFDParams} object containing mandatory physical definitions: geometric classification, Reynolds number ($Re$), fluid properties, flow regime, solver flags, and turbulence models. Furthermore, physical consistency is enforced via a post-hoc regular expression validation layer; if the LLM hallucinates a characteristic velocity fundamentally misaligned with the explicit $Re = U \cdot L_\text{char} / \nu$ relationship for the chosen fluid, the boundary conditions are automatically overridden and recalculated to maintain mathematical consistency.

\paragraph{Deterministic Solver Routing.}
Abstracting categorical solver selection away from the LLM eliminates a pervasive failure mode observed in early autonomous agents, where geometric keywords artificially bias the model toward inappropriate numerical regimes (e.g., erroneously selecting \texttt{simpleFoam} for transient vortex shedding). Instead, a deterministic routing algorithm maps the extracted physics flags (transient, compressible, thermal, multiphase) and Reynolds number against a rigorous heuristic table, successfully isolating one of seven validated OpenFOAM solvers.

\paragraph{Multi-Objective Reward Formulation.}
A critical component of the self-evolution pipeline is the autonomous scoring metric, denoted as $r \in [0,1]$. This scalar reward evaluates simulation fidelity across multiple physical and numerical axes, balancing mathematical convergence against mass conservation and boundary condition validity (Table~\ref{tab:reward}). The base reward is strictly additive up to an ideal score of $1.00$. Heuristic penalties are subsequently subtracted for violations of mesh quality (e.g., maximum non-orthogonality exceeding $70^\circ$), computational stagnation (residual plateaus), or excessive wall-clock execution times, with the final evaluation clipped to the $[0,1]$ interval. Simulations achieving an intermediate threshold of $r \geq 0.5$ are captured to populate the retry-context memory, while high-fidelity executions achieving $r \geq 0.65$ are immediately promoted to the curated training corpus for subsequent fine-tuning epochs.

\section{Dataset Curation and Trajectory Mining}
\label{sec:dataset}

To successfully transition general-purpose LLM reasoning into the highly specialized domain of computational fluid dynamics, AutoFOAM relies on a bifurcated data strategy: a statically curated foundational corpus to establish baseline syntax, and a dynamic trajectory log to power continuous, reinforcement-based alignment.

\begin{table}[h]
    \centering
    \caption{Reward weight for successful completion of each step in the workflow. }
    \label{tab:reward}
    \small
    \begin{tabular}{lrl}
        \toprule
        \textbf{Component} & \textbf{Weight} & \textbf{Trigger} \\
        \midrule
        Convergence (full)             & $+0.40$ & solver wrote final-time fields, residuals fell \\
        Convergence (partial)          & $+0.15$ & ran but did not converge \\
        Residual magnitude $< 10^{-4}$ & $+0.20$ & worst-converging field \\
        Residual magnitude $< 10^{-3}$ & $+0.10$ & worst-converging field \\
        Residual trend quality         & up to $+0.10$ & log-rate of decrease, no late spikes \\
        Mass conservation              & $+0.05$ & continuity error $< 10^{-3}$ \\
        Correct solver pick            & $+0.10$ & \texttt{solver = select\_solver(params)} \\
        Valid boundary conditions      & $+0.05$ & inlet \emph{and} outlet patches present \\
        \midrule
        Mesh quality penalty           & $-0.10$ & \texttt{checkMesh} non-orthogonality $> 70^\circ$ \\
        Slow runtime penalty           & $-0.10$ & wall-clock $> 300$~s \\
        Stagnation penalty             & $-0.05$ & residual plateau detected \\
        \bottomrule
    \end{tabular}
\end{table}
\subsection{Foundational Knowledge Corpus}
The baseline physical intuition and syntactic capability of AutoFOAM are derived from a highly curated, 402-row instruction-tuning dataset. Each record is structured as a multi-turn dialogue featuring an OpenFOAM expert persona. The model is trained to process a natural-language request, output a brief analytical summary, and generate complete, fenced Markdown representations of the required C++ dictionaries. This foundational corpus was synthesized by authoring 252 unique, edge-case-spanning prompts across all supported solver families and parametric geometries. These prompts were executed through early iterations of the agent, and only trajectories yielding a passing execution reward ($r \geq 0.5$) were preserved. This strict curation guarantees that the base model learns exclusively from mathematically converging and physically viable configurations.

\subsection{Dynamic Trajectory Capture and Preference Mining}
During live autonomous operations, the AutoFOAM pipeline persistently captures execution telemetry. Every highly successful simulation ($r \geq 0.65$) is appended to a dynamic capture dataset, forming a growing corpus that currently comprises 210 topologically de-duplicated scenarios. Simultaneously, the system maintains a comprehensive attempts ledger that records every failed or suboptimal execution, alongside its specific failure mode (e.g., continuity errors, diverging residuals) and the corresponding dictionary text. 

This attempt ledger is the critical mechanism driving Layer 4 of our self-evolution pipeline. By contrasting failed attempts with successful retries, the system autonomously mines explicit, on-policy preference pairs (chosen vs. rejected trajectories). These pairs are subsequently utilized for Direct Preference Optimization (DPO) \cite{rafailov2023direct}, allowing AutoFOAM to learn not just the correct syntax, but specifically how to navigate away from its own past hallucinations.

\subsection{Out-of-Distribution (OOD) Benchmark}
To rigorously evaluate the generalization capabilities of AutoFOAM and prevent overfitting to the synthetic training distribution, we constructed a strictly held-out evaluation set of 110 prompts. As detailed in Table~\ref{tab:ood_composition}, these prompts were authored using novel phrasing structures and idiosyncratic vocabulary that deliberately avoid the semantic patterns present in the foundational training catalog.
This benchmark serves as the absolute ground truth for assessing the agent's zero-shot physics reasoning and deterministic solver-routing capabilities.

\begin{table}[h]
    \centering
    \caption{Distribution of the 110-prompt OOD evaluation set, categorized by the canonically correct OpenFOAM solver family.}
    \label{tab:ood_composition}
    \begin{tabular}{lrr}
        \toprule
        \textbf{Expected Solver Profile} & \textbf{Count} & \textbf{Share} \\
        \midrule
        \texttt{simpleFoam} (steady incompressible)        & 74 & 67\% \\
        \texttt{rhoSimpleFoam} (steady compressible)       & 9  &  8\% \\
        \texttt{buoyantSimpleFoam} (steady buoyant)        & 7  &  6\% \\
        \texttt{interFoam} (multiphase VOF)                & 7  &  6\% \\
        \texttt{pimpleFoam} (transient turbulent)          & 6  &  5\% \\
        \texttt{icoFoam} (transient laminar)               & 4  &  4\% \\
        \texttt{rhoPimpleFoam} (transient compressible)    & 3  &  3\% \\
        \bottomrule
    \end{tabular}
\end{table}

\section{The Self-Evolution Pipeline}
\label{sec:evolve}

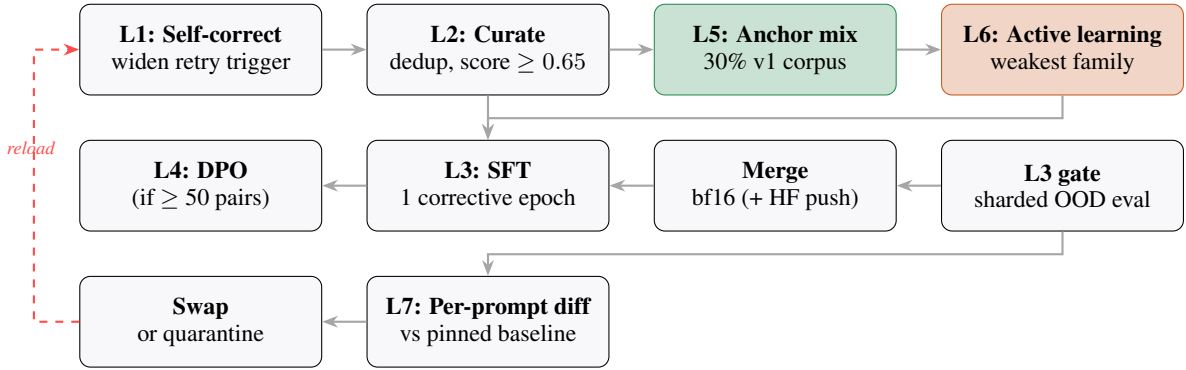
\begin{figure}[h]
		\centering
		\begin{tikzpicture}[
			every node/.style={font=\footnotesize},
			layer/.style={rectangle, rounded corners, draw, minimum width=32mm,
				minimum height=12mm, align=center, fill=ffcasebg,
				inner sep=2pt},
			anchorbox/.style={rectangle, rounded corners, draw, minimum width=32mm,
				minimum height=12mm, align=center, fill=ffanchor!25,
				draw=ffanchor, inner sep=2pt},
			activebox/.style={rectangle, rounded corners, draw, minimum width=32mm,
				minimum height=12mm, align=center, fill=ffactive!25,
				draw=ffactive, inner sep=2pt},
			arrow/.style={-{Stealth[scale=0.9]}, thick, gray!70},
			x=38mm, y=18mm,
			]
			\node[layer]     (n1) at (0,2) {\textbf{L1: Self-correct}\\widen retry trigger};
			\node[layer]     (n2) at (1,2) {\textbf{L2: Curate}\\dedup, score $\geq 0.65$};
			\node[anchorbox] (n5) at (2,2) {\textbf{L5: Anchor mix}\\30\% v1 corpus};
			\node[activebox] (n6) at (3,2) {\textbf{L6: Active learning}\\weakest family};
			\node[layer]     (l4) at (0,1) {\textbf{L4: DPO}\\(if $\geq$ 50 pairs)};
			\node[layer]     (l3) at (1,1) {\textbf{L3: SFT}\\1 corrective epoch};
			\node[layer]     (mg) at (2,1) {\textbf{Merge}\\bf16 (+ HF push)};
			\node[layer]     (gt) at (3,1) {\textbf{L3 gate}\\sharded OOD eval};
			\node[layer]     (sw) at (0,0) {\textbf{Swap}\\or quarantine};
			\node[layer]     (l7) at (1,0) {\textbf{L7: Per-prompt diff}\\vs pinned baseline};
			
			\draw[arrow] (n1) -- (n2);
			\draw[arrow] (n2) -- (n5);
			\draw[arrow] (n5) -- (n6);
			\draw[arrow] (n6.south) -- ++(0,-3mm) -| (l3.north);
			\draw[arrow] (n2.south) -- (l3.north);
			\draw[arrow] (gt) -- (mg);
			\draw[arrow] (mg) -- (l3);
			\draw[arrow] (l3) -- (l4);
			\draw[arrow] (gt.south) -- ++(0,-3mm) -| (l7.north);
			\draw[arrow] (l7) -- (sw);
			\draw[arrow, dashed, red!70] (sw.west) -- ++(-6mm,0) |- ($(n1.west)+(-6mm,0)$) -- (n1.west);
			\node[font=\scriptsize\itshape, red!70, anchor=east] at ($(n1.west)+(-2mm,-13mm)$) {reload};
		\end{tikzpicture}
		\caption{The seven-layer evolution protocol. The \textcolor{ffanchor}{green} module represents foundational memory retention (L5), and the \textcolor{ffactive}{orange} module signifies targeted active learning (L6). L7 functions as a strict regression firewall before promoting the new weights.}
		\label{fig:evolve}
	\end{figure}

A fundamental limitation of deploying LLMs as physical agents is their susceptibility to self-distillation collapse—a phenomenon in which training a model on its own-generated, slightly flawed output rapidly degrades its foundational capabilities \cite{shumailov2024curse}. To transition AutoFOAM from a static inference model into a continuously improving agent, we architected a rigorous, seven-layer self-evolution pipeline. Orchestrated autonomously, this loop explicitly digests production successes into positive reinforcement while converting failures into targeted preference-optimization gradients (Figure~\ref{fig:evolve}).

\subsection{The Seven-Layer Evolution Architecture}
The pipeline strictly regulates the flow of telemetry into the model weights through the following sequential protocols:

\begin{description}[leftmargin=12pt,style=nextline,topsep=4pt,itemsep=6pt]
    \item[\bf Layer 1: Autonomous In-Run Correction.] Upon detecting numerical divergence (e.g., $NaN$ residuals) or severe mass imbalance (continuity errors), the agent widens its context window to ingest the terminal OpenFOAM log output. This triggers an automatic re-prompting cycle, enriching the state context for a secondary parameter extraction attempt.

    \item[\bf Layer 2: Execution-Gated Curation.] The agent filters the live capture log, strictly promoting only high-fidelity executions achieving a reward threshold of $r \geq 0.65$. This acts as a primary filter against ingesting physically invalid topological hallucinations.
    
    \item[\bf Layer 3: Supervised Fine-Tuning (SFT) and Evaluation Gate.] The curated data is utilized for a single corrective QLoRA training epoch. Immediately following weight adaptation, the candidate model is subjected to an 8-way sharded OOD evaluation against a pinned performance baseline to ensure macro-level stability.
    
    \item[\bf Layer 4: Direct Preference Optimization (DPO).] Leveraging the mined trajectory logs, this layer contrasts a localized generation failure (rejected) against its eventually successful, Layer-1-corrected counterpart (chosen). DPO is applied to this on-policy preference pair, explicitly teaching the model to penalize the syntactic formulations that previously led to solver divergence \cite{rafailov2023direct}.
    
    \item[\bf Layer 5: Anchor Mixing.] To fiercely defend against catastrophic forgetting of core OpenFOAM syntax, the pipeline systematically injects a 30\% deterministic sample of the foundational training corpus into every dynamic retraining batch.
    
    \item[\bf Layer 6: Targeted Active Learning.] The agent analyzes the telemetry from the Layer 3 evaluation to identify the weakest-performing solver family. It then autonomously prompts itself to generate novel, parameter-varied seeds strictly within that domain, aggressively bolstering its weakest algorithmic flank.
    
    \item[\bf Layer 7: Granular Regression Auditing.] Before the newly trained adapter weights are permanently merged and pushed to deployment, the system performs a rigid, dictionary-level text diff against the pinned baseline. This traps minute structural regressions that aggregate scalar reward metrics might obscure.
\end{description}

\subsection{Surgical Recovery Protocols: Streams A and B}
When single-shot zero-shot generation fails to yield a mathematically convergent setup, AutoFOAM eschews a complete system reset in favor of two localized recovery protocols. \textbf{Stream A (Retrieval-Augmented Context)} queries a localized Chroma vector store \cite{chroma} containing validated OpenFOAM tutorials, injecting contextually analogous C++ dictionary structures into the prompt to guide recovery. Concurrently, \textbf{Stream B (Dictionary-Level Patching)} handles deterministic syntactic errors; if the solver yields a localized \texttt{FOAM FATAL} error concerning a specific file (e.g., an unsupported divergence scheme in \texttt{fvSchemes}), the agent bypasses the heavy \texttt{gmsh} meshing cycle entirely. Instead, it surgically targets and patches only the offending dictionary, escalating the generation failure into an iterative, highly efficient solver-feedback dialogue.

\section{Experiments and Results}
\label{sec:results}

We evaluate AutoFOAM across three primary axes: its capacity to autonomously synthesize valid physical meshes and boundaries (qualitative), its zero-shot generalization to novel linguistic prompts (quantitative), and the stability of its self-evolution pipeline against self-distillation collapse.

\subsection{Autonomous Case Setup and Flow Validation}
AutoFOAM natively supports seven core OpenFOAM solver families across 13 highly parametric \texttt{gmsh} geometry templates. These range from standard pedagogical benchmarks (e.g., lid-driven cavities, backward-facing steps) to complex industrial aerodynamic profiles (e.g., NACA airfoils, periodic hills, and 3D Ahmed bodies).

\begin{figure}[htbp]
    \centering
    
    \begin{subfigure}[b]{0.49\textwidth}
        \centering
        \includegraphics[width=\linewidth, height=5cm, keepaspectratio]{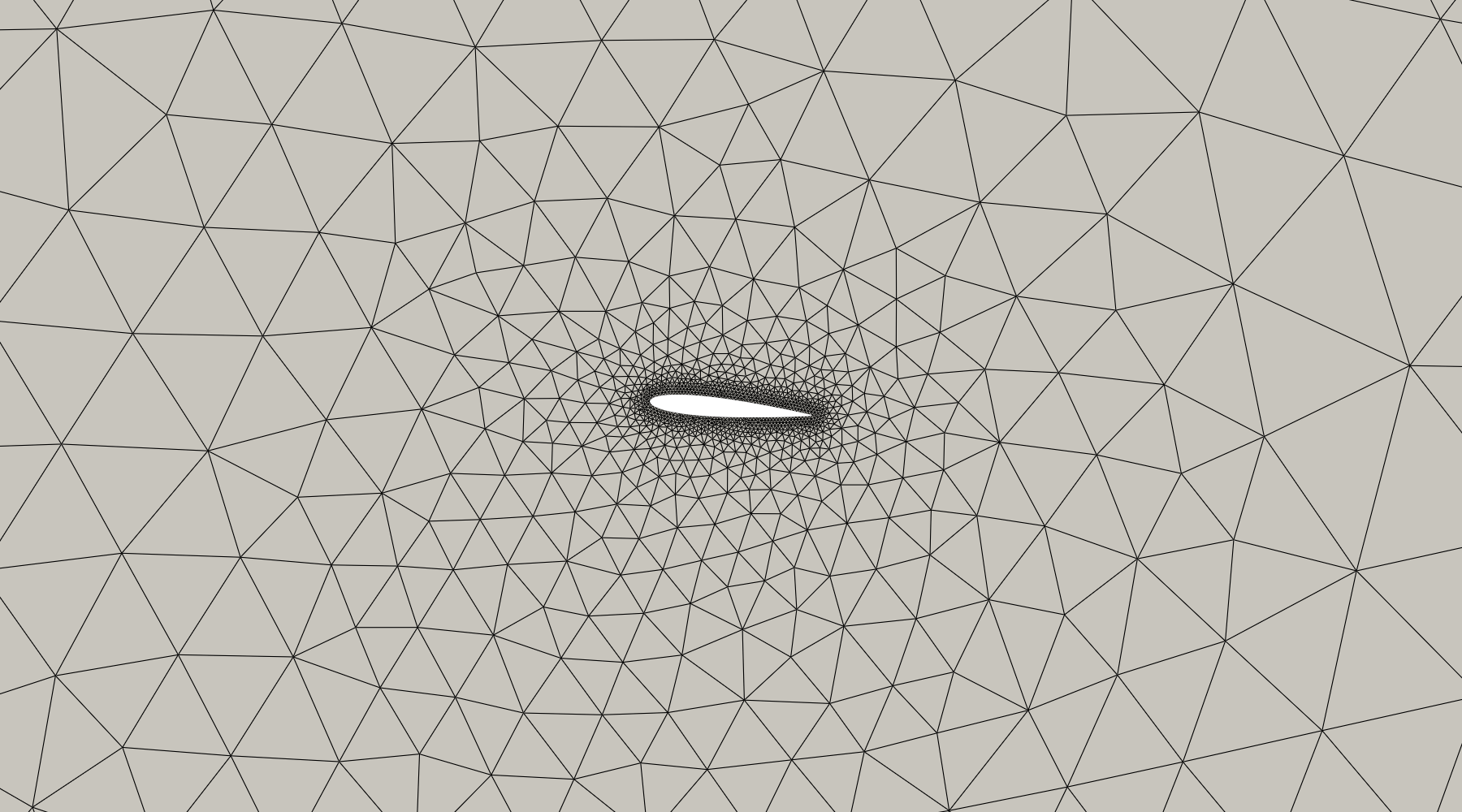}
        \caption{Zoomed view: NACA-4-digit airfoil.}
    \end{subfigure}\hfill
    \begin{subfigure}[b]{0.49\textwidth}
        \centering
        \includegraphics[width=\linewidth, height=5cm, keepaspectratio]{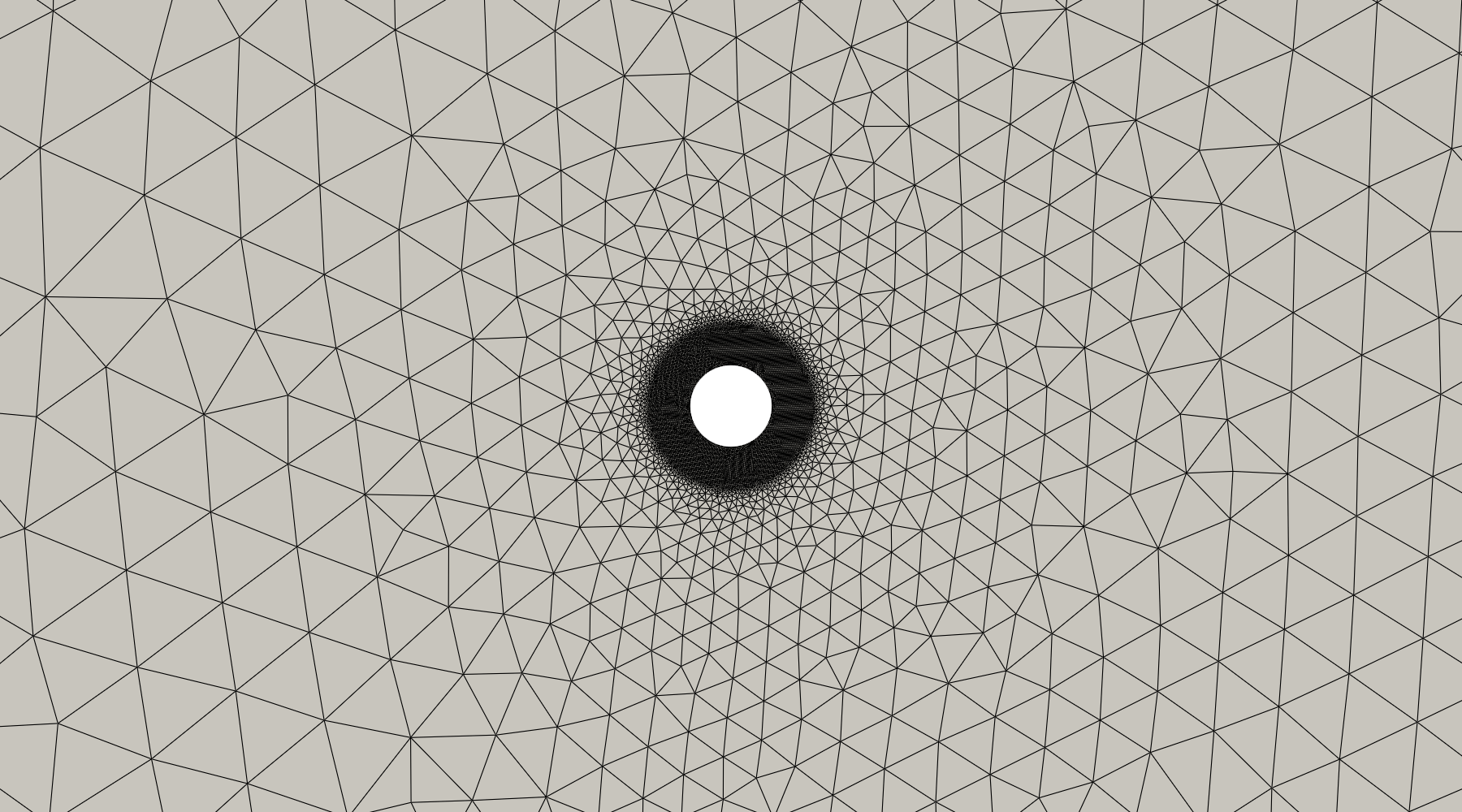}
        \caption{Zoomed view: Cylinder in cross-flow.}
    \end{subfigure}
    
    \vspace{1.5em} 
    
    \begin{subfigure}[b]{0.49\textwidth}
        \centering
        \includegraphics[width=\linewidth, height=5cm, keepaspectratio]{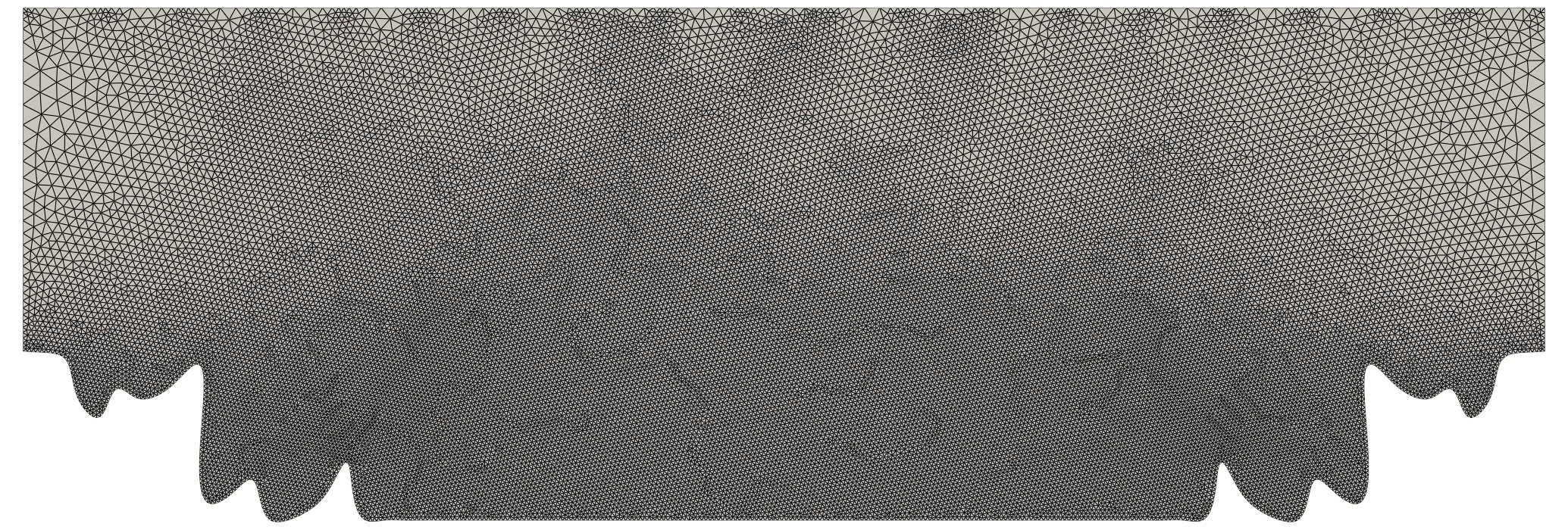}
        \caption{Periodic-hill-Mellen Polynomial.}
    \end{subfigure}\hfill
    \begin{subfigure}[b]{0.49\textwidth}
        \centering
        \includegraphics[width=\linewidth, height=5cm, keepaspectratio]{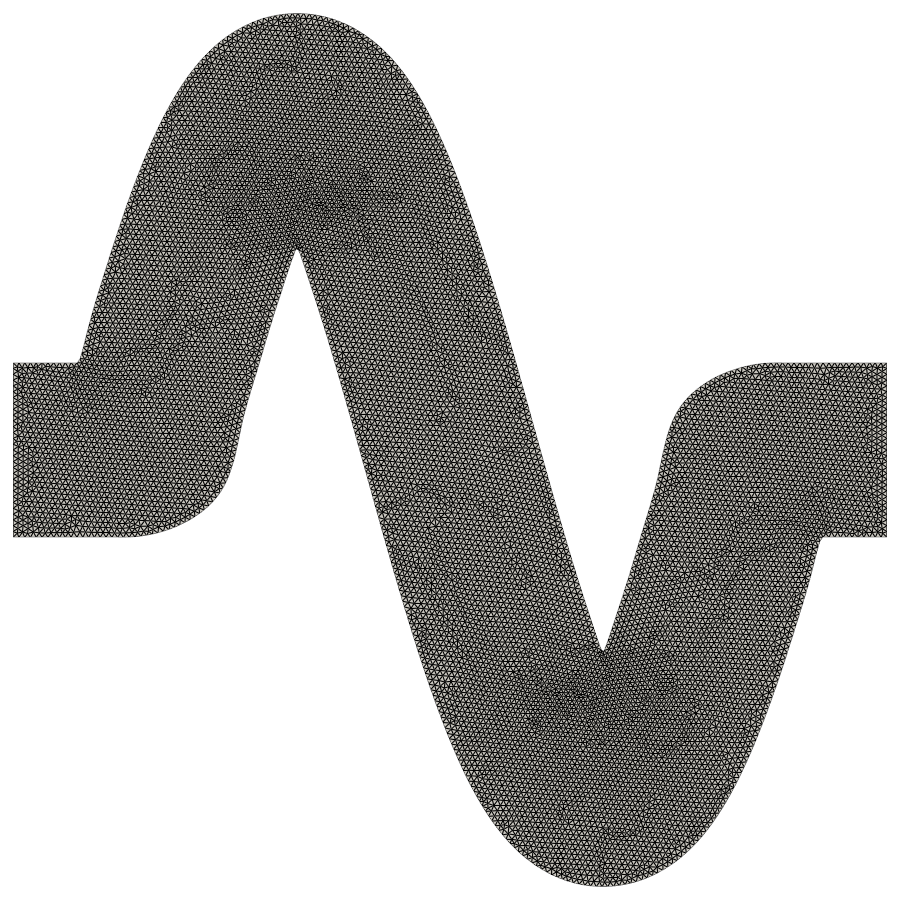}
        \caption{Flow inside S-bend}
    \end{subfigure}
    
    \caption{Mesh generated by AutoFOAM}
    \label{fig:meshes}
\end{figure}

\begin{figure}[htbp]
    \centering
    
    \begin{subfigure}[b]{0.49\textwidth}
        \centering
        \includegraphics[width=\linewidth, height=5cm, keepaspectratio]{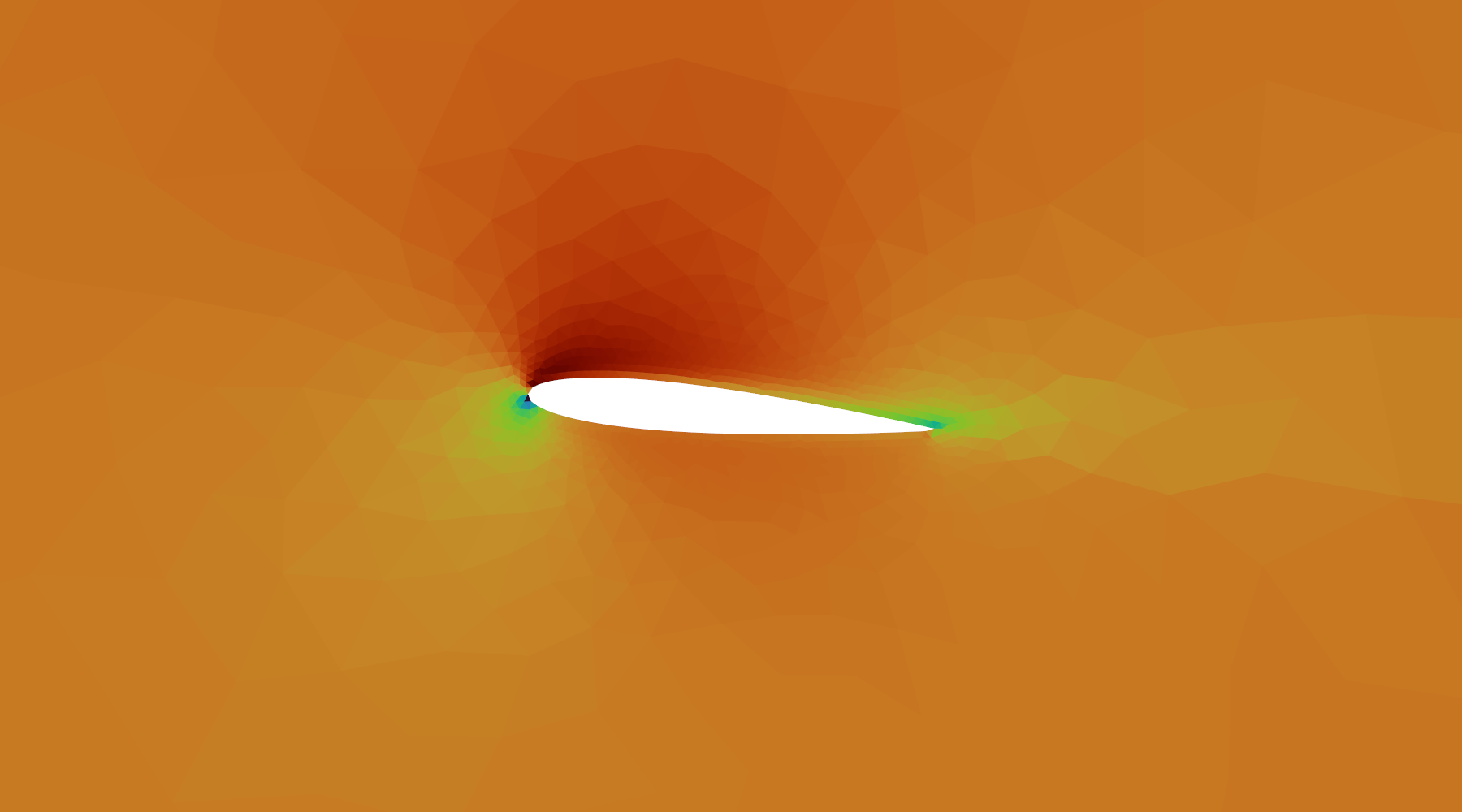}
        \caption{NACA-4-digit airfoil.}
    \end{subfigure}\hfill
    \begin{subfigure}[b]{0.49\textwidth}
        \centering
        \includegraphics[width=\linewidth, height=5cm, keepaspectratio]{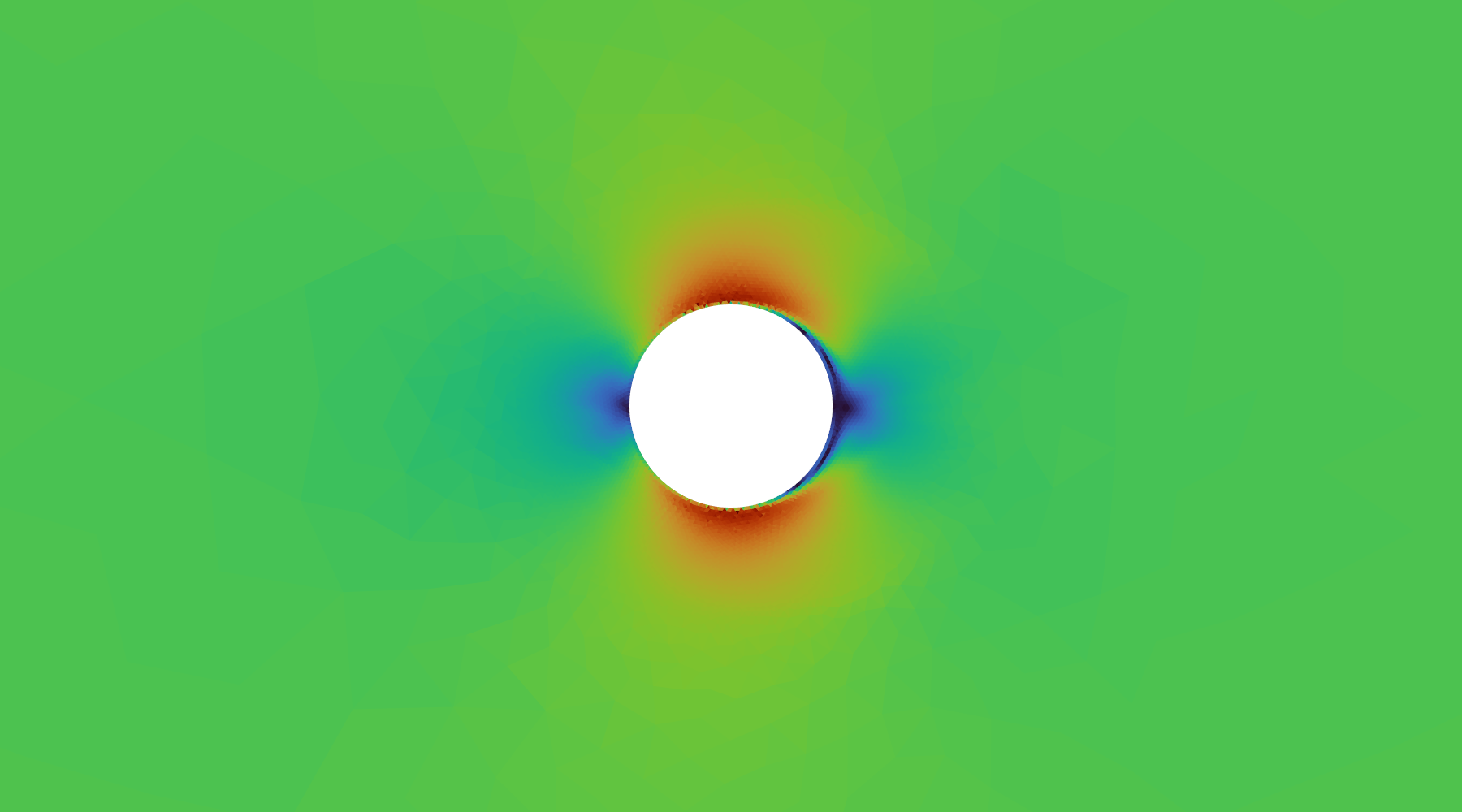}
        \caption{Cylinder in cross-flow.}
    \end{subfigure}
    
    \vspace{1.5em} 
    
    \begin{subfigure}[b]{0.49\textwidth}
        \centering
        \includegraphics[width=\linewidth, height=5cm, keepaspectratio]{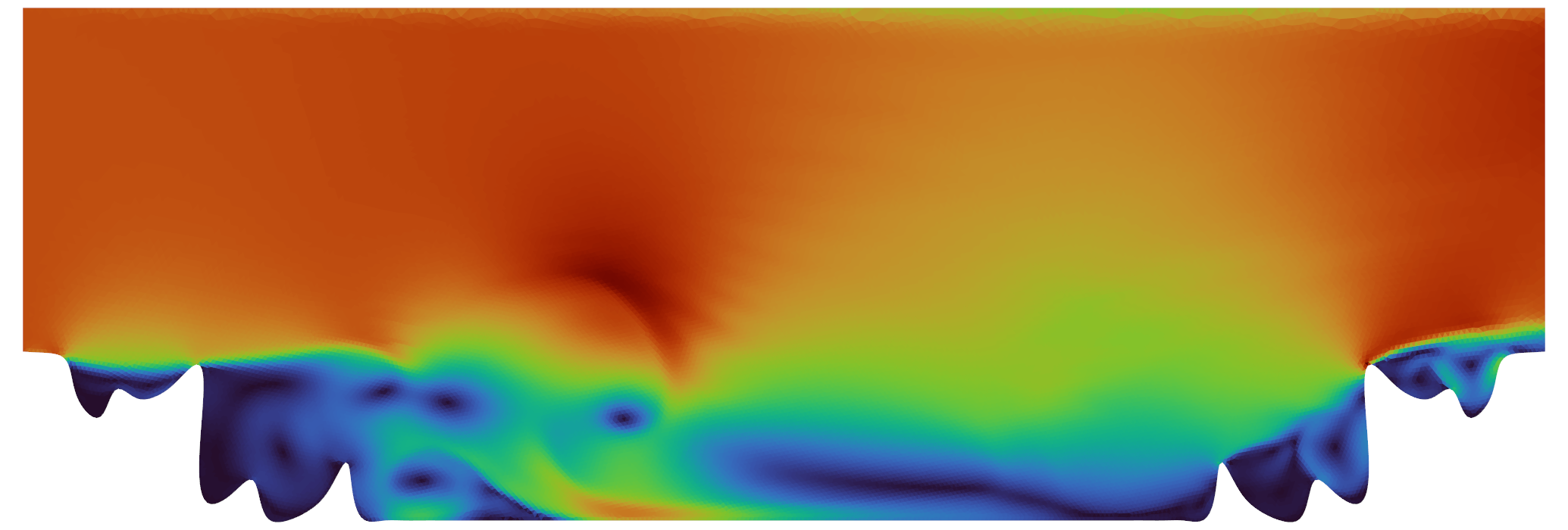}
        \caption{Periodic-hill-Mellen Polynomial.}
    \end{subfigure}\hfill
    \begin{subfigure}[b]{0.49\textwidth}
        \centering
        \includegraphics[width=\linewidth, height=5cm, keepaspectratio]{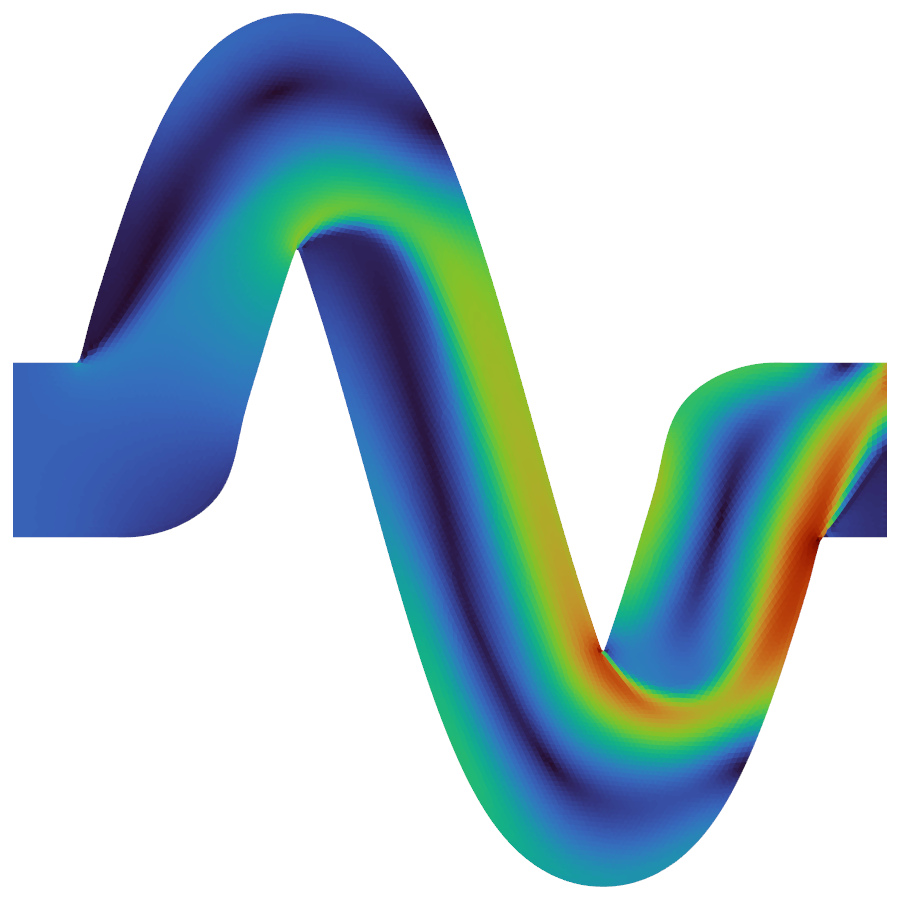}
        \caption{Flow inside S-bend}
    \end{subfigure}
    
    \caption{Velocity contours generated by AutoFOAM}
    \label{fig:flows}
\end{figure}

To visually confirm numerical convergence and physical viability, Figure~\ref{fig:flows} illustrates velocity-magnitude contours generated entirely autonomously from single-sentence inputs. Without any manual intervention, the agent successfully enforces appropriate \texttt{polyMesh} boundary naming conventions, constructs structurally sound grids (Figure~\ref{fig:meshes}), and configures proper turbulence conditions (e.g., $k$-$\omega$ SST parameters derived from the extracted Reynolds number), resulting in expected physical phenomena such as steady attached wakes and suction-side acceleration.

\subsection{Out-of-Distribution Zero-Shot Generalization}
The absolute efficacy of the autonomous pipeline was benchmarked using a strictly held-out set of 110 out-of-distribution (OOD) prompts, executed on an 8-way sharded cluster of NVIDIA H100 GPUs. The evaluation was restricted to solver families and geometries supported by AutoFOAM, while varying the geometric parameters and physical operating conditions to ensure true OOD generalization. In parallel, we also assessed AutoFOAM’s capability to autonomously select the appropriate solver based solely on user-described physics scenarios, assuming no prior user expertise in solver selection. \textit{Please note that the accuracy of AutoFOAM will degrade if we ask it to perform tasks outside the solver or the geometry supported by the current agent}.
As detailed in Table~\ref{tab:headline}, the fine-tuned agent successfully executed 100\% of the evaluation cases end-to-end without a single \texttt{FOAM FATAL} failure.

\begin{table}[h]
    \centering
    \caption{OOD Evaluation Metrics.}
    \label{tab:headline}
    \begin{tabular}{lr}
        \toprule
        \textbf{Metric} & \textbf{Value} \\
        \midrule
        End-to-end execution without FOAM FATAL & $110/110$ ($100.0\%$) \\
        Solver routing exact match  & $106/110$ ($96.4\%$) \\
        Mean execution reward score ($r$)  & $0.64$ \\
        Geometry templates successfully exercised & 13 / 13 \\
        Solver families successfully exercised   & 7 / 7 \\
        \bottomrule
    \end{tabular}
\end{table}

Furthermore, the agent demonstrated robust physical reasoning capabilities, achieving an exact match rate of $96.4\%$ on deterministic solver routing. A manual review of the four mismatches (Table~\ref{tab:prompt_examples}) reveals that they occurred exclusively in mathematically ambiguous boundary regimes, specifically, low-Reynolds unsteady cylinder configurations, where the agent selected the laminar \texttt{icoFoam} solver instead of the strictly mandated \texttt{pimpleFoam}. From a computational physics standpoint, this substitution remains highly defensible and does not indicate a breakdown in the model's semantic understanding.

\begin{table}[h]
    \centering
    \caption{Representative OOD prompts illustrating the agent's semantic understanding and correct translation to physical solvers.}
    \label{tab:prompt_examples}
    \small
    \begin{tabularx}{\linewidth}{lXll}
        \toprule
        \textbf{Benchmark Tag} & \textbf{Prompt} & \textbf{Predicted Solver} & \textbf{Match} \\
        \midrule
        \texttt{cav\_re150\_water}    & \emph{``Cavity flow Re=150 in water, 0.4~m square''} & \texttt{simpleFoam} & \checkmark \\
        \texttt{pipe\_water\_lam}     & \emph{``Laminar pipe flow Re=500 diameter 0.02~m, water''} & \texttt{simpleFoam} & \checkmark \\
        \texttt{airfoil\_re1m\_aoa5}  & \emph{``NACA0012 airfoil at Re=$10^6$, 5$^\circ$ angle of attack''} & \texttt{simpleFoam} & \checkmark \\
        \texttt{inter\_dam\_tall}     & \emph{``Tall dam break, 2~m wide, 5~m water column collapse''} & \texttt{interFoam} & \checkmark \\
        \texttt{cdnoz\_ma04}          & \emph{``Convergent-divergent nozzle with Mach 0.4 throat, air''} & \texttt{rhoSimpleFoam} & \checkmark \\
        \texttt{cyl\_re250\_unsteady} & \emph{``Transient cylinder Re=250 in water, D=0.1~m''} & \texttt{icoFoam} & $\times$ \\
        \bottomrule
    \end{tabularx}
\end{table}

\section{Scope and Limitations}
\label{sec:limitations}

While the current AutoFOAM framework demonstrates robust zero-shot generalization and stable self-improvement, several operational envelopes remain out of scope for this deployment. Geometrically, the agent is constrained to the 13 defined \texttt{gmsh} parametric templates; raw ingestion of complex industrial CAD formats (e.g., raw STEP or IGES files) requiring autonomous topology repair and unstructured tetrahedral meshing is not yet supported. Physically, the numerical policy is currently limited to standard RANS turbulence modeling and laminar flows. Advanced formulations such as hybrid RANS-LES, dynamic adaptive mesh refinement (AMR), and conjugate heat transfer algorithms require significantly more complex dictionary configurations that the current foundational corpus does not yet cover.

\paragraph{The Physics-Validation Gap.}
The most critical limitation of the current architecture lies within the multi-variate reward function (Section \ref{sec:arch}). Presently, the scoring system heavily prioritizes mathematical convergence (residual decay) and mass conservation (continuity). However, a simulation can be mathematically stable while remaining physically flawed (e.g., a converged solution with an entirely incorrect drag coefficient due to an improperly estimated reference velocity). Because the agent lacks an explicit cross-referencing module to validate its outputs against empirical data or established literature (such as comparing lid-cavity centreline profiles against benchmark studies \cite{ghia1982}), physically inaccurate but numerically converged cases can inadvertently pass the Layer 2 curation gate. This "physics-validation gap" risks starving the Layer 4 DPO loop of vital negative preference signals. Implementing a strict, external \texttt{physics validator} to independently verify the integrity of the flow field is the primary directive for future iterations of this work.

\section{Conclusion}
\label{sec:conclusion}

In this work, we introduced AutoFOAM, a fully autonomous LLM-driven OpenFOAM agent capable of bridging the gap between natural language intent and rigorous computational fluid dynamics case execution. By anchoring a fine-tuned 14-billion-parameter Qwen2.5-Coder model with strict deterministic constraints, the system successfully eliminates the cognitive and syntactic barriers associated with manual CFD configuration. AutoFOAM achieves a 100\% execution success rate and a 96.4\% solver-routing precision on novel, out-of-distribution prompts. 
Crucially, this work moves beyond static inference by implementing a rigorous, seven-layer self-evolution loop. By strategically combining targeted Active Learning, Direct Preference Optimization (DPO) on mined failure trajectories, and systematic Anchor Mixing, we provide a blueprint for safely scaling agent capabilities without succumbing to self-distillation model collapse. The resulting architecture serves not only as a highly practical assistant for accelerating engineering workflows, but also as an open-source testbed for researching continual, reinforcement-driven learning within strict physical constraints.

\section*{Artefacts and Reproducibility}
To foster reproducible research in scientific machine learning, all model weights, codebases, training scripts, and foundational datasets are released openly under the Apache 2.0 license:
\begin{itemize}
    \item \textbf{Codebase:} \href{https://github.com/AGN000/AutoFOAM}{\texttt{https://github.com/AGN000/AutoFOAM}}
    \item \textbf{Dataset:} \href{https://github.com/AGN000/FoamAgentCases}{\texttt{github.com/AGN000/FoamAgentCases}}
    \item \textbf{Model Weights:} \href{https://huggingface.co/arungovindneelan/foam-cfd-unified-14b}{\texttt{huggingface.co/arungovindneelan/foam-cfd-unified-14b}}
\end{itemize}
\section*{Acknowledgment}

The authors acknowledge the support of the \textbf{NVIDIA Inception} startup program. 
The program provided valuable access to the NVIDIA ecosystem, technical guidance, 
and computing resources that supported the development and experimentation of 
\textit{AutoFOAM}, an autonomous OpenFOAM agent framework for AI-driven CFD workflows.
\bibliographystyle{unsrt}
\bibliography{references}

\end{document}